# Detecting Generic Music Features with Single Layer Feedforward Network using Unsupervised Hebbian Computation

Sourav Das
*National Institute of Electronics & Information Technology, Kolkata, India*

Anup Kumar Kolya
*Department of Computer Science, RCC Institute of Information Technology, Kolkata, India*

**ABSTRACT**

*With the ever-increasing number of digital music and vast music track features through popular online music streaming software and apps, feature recognition using the neural network is being used for experimentation to produce a wide range of results across a variety of experiments recently. Through this work, the authors extract information on such features from a popular open-source music corpus and explored new recognition techniques, by applying unsupervised Hebbian learning techniques on their single-layer neural network using the same dataset. The authors show the detailed empirical findings to simulate how such an algorithm can help a single layer feedforward network in training for music feature learning as patterns. The unsupervised training algorithm enhances their proposed neural network to achieve an accuracy of 90.36% for successful music feature detection. For comparative analysis against similar tasks, authors put their results with the likes of several previous benchmark works. They further discuss the limitations and thorough error analysis of their work. The authors hope to discover and gather new information about this particular classification technique and its performance, and further understand future potential directions and prospects that could improve the art of computational music feature recognition.*

## INTRODUCTION

Music features are characteristic classifications that are used to distinguish between different genres of music. Also, each genre differs from other genres in certain musical features. In the pre-computational intelligence era, music feature categorization has traditionally been performed manually, mostly due to the lack of modern human-computer interaction concept, and obviously for the lack of enough computational processing abilities of the computers. However, with the ever-increasing number of digital music and vast features, feature recognition using the neural network is being producing a wide range of results across a variety of experiments recently. Nowadays, music feature classification is a popular topic for research, particularly in the fields of Music Information Retrieval (MIR) and Neural Network. Since Artificial Neural Network with a variety of learning methods have been introduced in this domain, there have been many research works done on methods for classifying music features according to the diversity of their features. The human brain computes in a distinctly different way rather than conventional computers. Keeping this concept in mind, the research work on Artificial Neural Network has been motivated. The main ideology of a Neural Network is to represent a linear or nonlinear and parallel computing architecture. Researchers have also suggested methods for music feature classification using multiple feature vectors and pattern recognition ensemble approach. In figure 1, the authors represent the process blocks of traditional music classification techniques utilizing neural networks.

Music segments are also decomposed according to time segments obtained from the beginning, middle and end parts of the original music signal (time-decomposition), alongside with distinguished musical features such as pitch, intensity, beat, drop and so on, which is unlikely to be the same for even two distinct music files. In the present scenario, one of the biggest bottlenecks in many Music Information Retrieval (MIR) tasks is the access to large amounts of music data and their features, in particular to audio features extracted from commercial music recordings (Porter et al., 2015). However as of now, at least hypothetically these challenges can also be addressed by using a multilayer neural network with more than one hidden layer within it, and with backpropagation feedback signaling. Previous work in this topic represented a robust hypothesis to show how the music genre recognition can be done from distinct musical features using a single-layered feedforward neural network (Das & Kolya, 2019).

In this paper, the authors carry forward the problem domain further, to propose an empirical technique of successfully tackling the feature diversity of music. They present a simple yet featureful work to show the performance of a shallow neural network or single-layered neural network on an already established music feature corpus with unsupervised learning. They focus on the empirical computation that takes place while the network learns basic but distinct music properties such as beat, Fourier signal transmission, Mel-frequency signal, and finally pitch of the tunes. These music feature recognitions can be done from features converted as distinct binary inputs with the Hebbian learning technique, which is at the core of the proposed approach. The authors are here interested in observing and presenting the intricate differences like such signals when they pass through a single layer network following and unsupervised learning pattern. They further examine and observe each music features passing to their respective perceptrons and behavioral expression through the activation summarizations. Finally, they visually observe and note the performance of the proposed method to classify the errors. The authors achieve an overall accuracy of 90.36% in the validation phase. From that, they also perform a comparative evaluation with such state-of-the-art works in a similar domain and propose the paths for future enhancements.

*Figure 1. Ground-level music classification using neural networks*

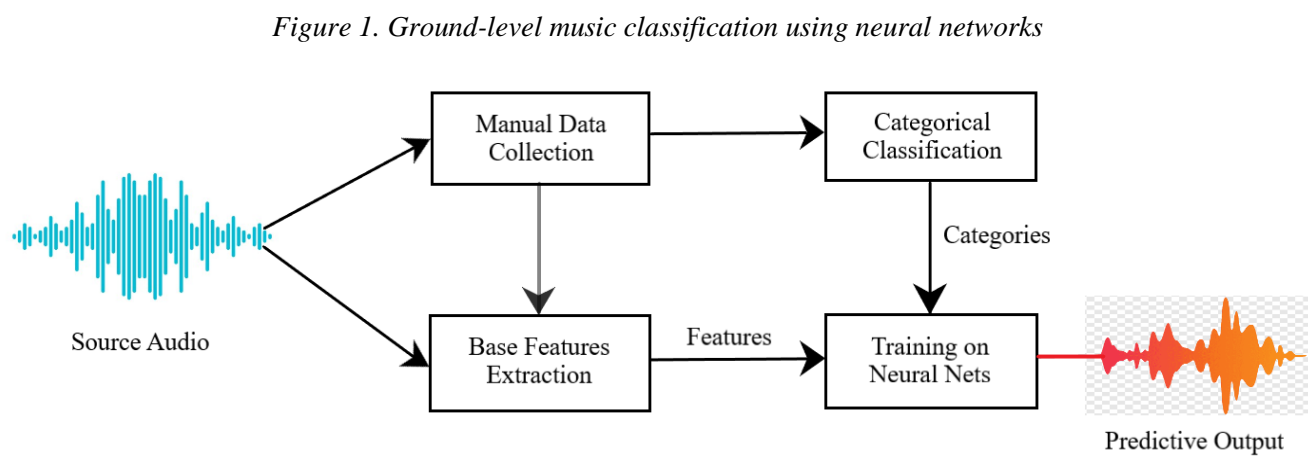

Main contributions of this paper are:

1. The authors present the novel approach for phase by phase empirical computation of how a single layer feedforward network learns sophisticated music features.
2. The learning technique is more emphasized here as it is unsupervised; there are no predefined patterns for training supervision.
3. They show perceptron level behavioral tuning and observations of the signal, which is self-explanatory for the network's output generation performance.

The rest of this paper is organized as follows:

In Section 2, the authors discuss some relevant works from a similar domain. In Section 3, they state the motivation behind their proposed approach. Section 4 elaborates about the dataset for their experimentation and feature engineering upon that. The authors represent the computational architecture of a single layer feedforward neural network in Section 5. In Section 6, they discuss the unsupervised Hebbian algorithmic computation for their proposed approach. The authors mention their experimental setup and neural network tuning in Section 7. Section 8 deals with the training details, whereas Section 8 represents the observational results. The authors also perform detailed comparative analysis with the other works for a similar task in this section. The authors analyze the runtime complexities for the execution of their proposed method in Section 10. The authors discuss the limitations of their work and future directions from this work in section 11. In Section 12, they conclude their work.

**BACKGROUND WORKS**

Related work contains a brief but orderly description of research works carried out on this domain so far. To understand the evolving trend of content-based feature recognition in Music Information Retrieval using Neural Network over the past years, the authors have carried out the literature survey thoroughly from the early 2000s to the most recent and relevant works. Each research work has made a positive impact on their way of approach the matter and objectifying new scope and possibilities.

(Tzanetakis et. al., 2002) stated that manual categorization of music is commonly used to classify and describe music. But the increment of digital music on the internet over time has led to the inevitability of Music Information Retrieval. Through their research work, they explored the algorithms for automatic genre classification. They also proposed a set of features for representing the music surface and rhythmic structure of audio signals. The performance of the training dataset represent-ed by them was measured and evaluated by training statistical pattern recognition classifiers using audio collections collected from compact disks, radio, and the web relevant at that time.

In another research work, 109 musical features were extracted from music playback and recordings and were classified according to their genres (McKay, 2004). The features were based on the distinct characteristics of the music files that were used for the experiment. The classification experiment was carried out using different sets of features for different hierarchical levels, which were determined by using genetic algorithms.

Pattern recognition combined with a vector-based approach has also been made for music genre classification problem (Silla et al., 2008). The researchers labeled music genre classification by binary classifiers and finally concluded the result by merging all the results gathered from each test case. Each audio file was decomposed based on time segments such as beginning part, middle part, and the end part of each song. Besides using conventional Machine Learning Algorithms, authors have also implemented Support Vector Machines (SVM) and Multilayer Perceptron Neural Networks.

Deep Belief Network (DBN) can also be used to extract features from audio files. A Deep Belief Network is a Multilayer Feedforward Neural Network, with multiple (deep) layers of hidden layers between the input and output layer. In a work, (Hamel & Eck, 2010) trained the activation functions of this network with a non-linear Support Vector Machine. This training eventually led them to solve the music genre identification problem up to an extent with an accuracy of 84.3%.

(Dieleman et. al., 2011) developed a Convolutional Neural Network with an unsupervised learning method that was trained to produce results such as artist recognition, genre recognition, etc. They experimented with the 'Million Song Dataset' (Bertin-Mahieux et al., 2011). The audio dataset was fragmented and labeled, and the multi-layer perceptrons were trained with those labels for each task. Henceforth, these researchers observed that their system gave an improved accuracy for both genre and artist recognition. They also stated that the unsupervised method of learning ensured comparatively a more rapid convergence of produced accurate results.

Tacit knowledge earned by a neural network can also be used to extract the common features from music (Humphrey et al., 2012). Rather than only audio-based music information retrieval, Eric J. Humphrey and his fellow researchers approached implementing the deep signal processing architecture. Through their work, the researchers observed that improved results about music information can be gathered through breaking a large network into smaller and simpler fragments or parts. They have also observed that applying the learning methods in flexible machines can make a positive impact on musical feature extraction.

It is also observed that using Deep Neural Network for audio event classification can produce better results (Kons et al., 2013). It is said to perform better than the Support Vector Machine for the same task. The neural network was trained with the Restricted Boltzmann Machine, which itself is a Genetic Algorithm's selection method. Boltzman Machine is used to produce a stable distribution of all available sample test population.

A drawback of applying neural networks for Music Information Retrieval can be relatively larger training time of the network, concerning the training process often gets stuck with local minima, i.e. considering the current best result as the universally best result in a widely distributed neural network architecture, when it is not. (Sigtia & Dixon, 2014) experimented with the ways to improve the learning capacity of a neural network in reduced time with the help of Rectified Linear Units (ReLU) as the activation functions, and expansion-conjugation methods such as Hessian-Free Optimization. (Dai et. al., 2015) stated that, since musical genres do not consist of the "multi-lingual" feature, hence they implemented the 'nearest neighbor algorithm' to label the large musical database again afterward classification. They trained a multi deep neural network (DNN) using a large musical database, so the DNN can learn quickly and transfer the musical files to a similar but smaller dataset, which is the target dataset. Finally, the authors evaluated the performance using a benchmark for many such popular music databases.

In another work from the same year, (Sharma et. al., 2015) presented a work to classify audio songs based on their music pattern to finally finding and sorting out the music clips based on the listener's taste. The research finding is said to be helpful in indexing and accessing the particular music file based on the listener's state. The authors demonstrated seven categories of listening moods for the work such as devotional, energetic, folk, happy, pleasant, sad, and, sleepy. They also considered some harmonic and listening specific features such as jitter and shimmer along with the Mel-frequency cepstral coefficients (MFCCs). Statistical values of pitch such as min, max, mean, and standard deviation are computed and added to the MFCCs. The researchers worked on a feedforward backpropagation neural network (BPNN) as the classifier due to its efficiency in mapping the nonlinear relations. They achieved an accuracy of 82 % on an average for 105 testing audio files. (Jeong & Lee, 2016) explained the framework for a mundane audio feature learning system using the deep neural network, leading it to the implementation of music genre classification. Also, they observed the conventional spectral feature learning framework and modified it with the cepstral modulation spectrum domain for better performance, which already has been applied previously in much successful speech or music feature learning experiments.

(Schindler et al., 2016) took an approach to extract subject-specific music-related visual information from music videos. They introduced the Music Video Dataset, with dedicated evaluation tasks for finding out the information from the music videos that are aligned to current Music Information Retrieval tasks. Simultaneously from this dataset, they also provided evaluations of conventional low-level image processing and particular music affect-related features for individual clips to provide an overview of the expressiveness of primary visual properties such as color, illumination, and contrasts. With that, they parallelly introduced a high-level approach based on the previously achieved visual concept detection to facilitate visual stereotypes. This approach said to be decomposing the semantic content of music video frames develops the meaning of a video into concrete concepts. These concepts are detected using convolutional neural networks and their frequency distributions as semantic descriptions for a music video. Finally, they observed highly significant performance improvements by augmenting audio-based approaches through the introduced visual approach for a meaningful dissertation of the music video clips.

(Choi et. al., 2017) introduced a convolutional neural network for music tagging by extracting the local musical features and summarizing those extracted features. Furthermore, they compared the performance of the Convolutional recurrent neural network that they used, with the performance of a general convolutional neural net-work concerning input parameters and training time per feature. They observed and presented that the CRNN they proposed, actually delivers better performance due to the heterogenicity of this network in music feature extraction and summarization.

(Liu et. al., 2018) applied a convolution neural network (CNN) to identify emotion from songs using time and frequency information by classifying spectrogram. (Nasrullah & Zhao, 2018) took temporal structure as a feature for artist classification using a convolutional recurrent neural network (CRNN) and experimented on the artist20 music dataset. (Lv et al., 2018) used a support vector machine and convolutional neural network for music emotion classification such as sad, exciting, serene, and happy using music feature set as input. (Ghosal & Kolekar, 2018) applied an ensemble technique of convolutional neural network and long short-term memory (CNN LSTM) and transfer learning model for music genre recognition. They took a rhythmic and diverse set of spectral features for their mode. (Huang et. al., 2019) applied a deep convolutional neural network with feature fusion to identify pitch, duration, and symbol categories from an image. (Pérez-Marcos at al., 2019), they used social media data as multi-agent system information for music feature extraction.

## MOTIVATION

The literature survey leads us to identify possible research areas that can be extended further. The aspects which can be carried out for further research are:

1. The primitive motivation for the proposed approach is to enhance the natural sound-enhancing capabilities of machines and overall reflecting the growth for the betterment of human-computer interaction. It is already established that natural language processing can be related to music information retrieval (Amiri, 2016). However, it is yet unchallenged that how will the neural network perform in enhancing the learning capacity of machines by classifying natural and distinct features of music files.
2. The neurons of each input unit from the input layer of a neural network accept bipolar (i.e. 0 or 1) values as the primary training input. Hence, labeling different genres of music using binary values (e.g. 001, 010, 011, etc.) is needed to build the input training patterns for a neural network. It would be interesting to observe the learning and thereafter recognition performance of a neural network trained with such perquisite values.

## DATA FOR EXPERIMENT

The widely popular dataset for music genre or musical feature recognition is the GTZAN dataset (Sturm, 2013), which was originally developed by George Tzanetakis. It is a dataset of size 1.5 Gb, consisting of 100 music examples of 10 different musical gen-res consisting of 30 seconds sample for each music, such as Hip Hop, Rock, Metal, Jazz. Another available and versatile dataset is the Million Song Dataset, which has been discussed previously (Bertin-Mahieux et. al., 2011). The Million Song Dataset contains 280 Gb of data with 1,000,000 songs per file, 44,745 unique artists, etc.

The authors made the split validation of the aforesaid datasets in a ratio of 0.66 and 0.34, i.e. 66% of the dataset was implemented for training purpose, while the rest 34% of data is applied and will be applied further for their extensive work for testing purpose once the training epochs are completed. They discuss the feature extraction 66% of training data, for better training purpose of the proposed method.

   1.   Feature Extraction

The feature extraction from the previously discussed data comprises some sequential phases. The authors approach for selecting such features as properties utilizing the simple Python music and speech library. The authors mention them in consecutive orders.

1. The authors first aim for the zero-crossing rate, which is the rate of significant changes with a signal. It is determined when the rate at which the signal changes from positive to negative and vice versa. This feature has already been used heavily in music information retrieval and features extraction from corpora. It usually has higher values for highly percussive sounds like those in metal and rock, and lower values for normalized timid sounds like blues and jazz.
2. Next, they aim for filtering the spectral centroids for the soundtracks from the corpus. It precisely indicates the center of weight of a sound, to further calculate the weighted mean of the frequencies generated in the neighboring signals present surrounding the centroid itself. If the frequencies in music are the same throughout then spectral centroid for a given period, it would be around a center and if there are high frequencies at the end of sound then the centroid would be towards its end.
3. Finally, they aim to filter the cepstral coefficients extraction for Mel-frequency generation. This method helps to extract whether an audio signal and is used majorly whenever working on audio signals. The Mel-frequency cepstral coefficients (MFCCs) of a signal are a small set of features, which concisely describe the overall shape of a spectral coverage. By printing the shape of MFCC later, they can get the direct relationship between how many numbers are calculated on how many frames. The first parameter represents the number of MFCCs calculated and another value represents several frames available.

The authors represent the feature extraction process in Figure 2.

*Figure 2. Feature Extraction Model from Music Corpus*

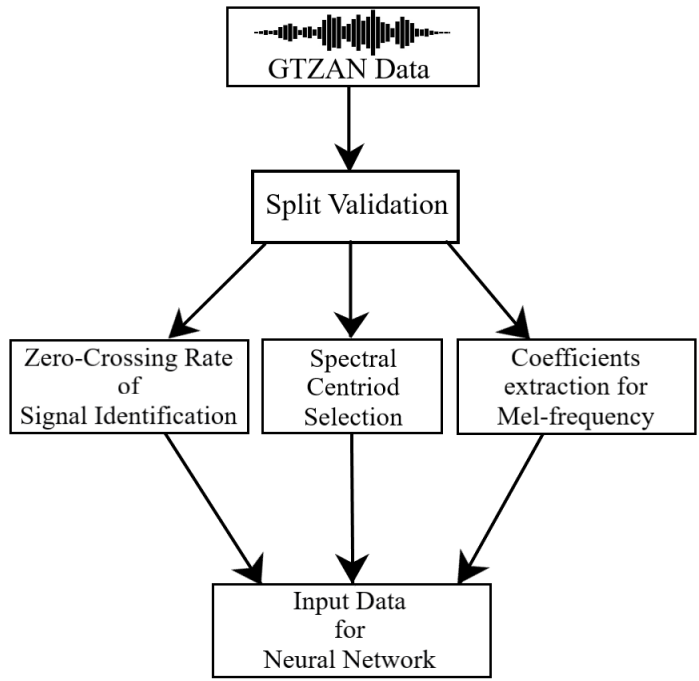

## LAYOUT OF A FEEDFORWARD NEURAL NETWORK

As the authors have mentioned in Section 1, there are several classes of Neural Network, such as Single Layer Feedforward Neural Network, Multi-Layer Feedforward Neural Net-work, Recurrent Network, etc. A single layer network comprises two layers, one input, and one output layer. Whereas, multilayer network is made up of multiple layers, i.e. input, output, and one or more intermediary layers called hidden layers.

For their work, the authors have followed the simplistic approach of Single Layer Feedforward Neural Network. In SLNN, only the output layer can perform the computation, hence it is called the single layer. Besides, the synaptic links carrying the weights (Siegelmann & Sontag, 1994), connect every input neuron to the output neurons, but not the reverse. Hence it is called the feedforward network.

The authors show the diagram of a Single Layer Feedforward Neural Network in Figure 3.

*Figure 3. Single-layer feedforward neural network*

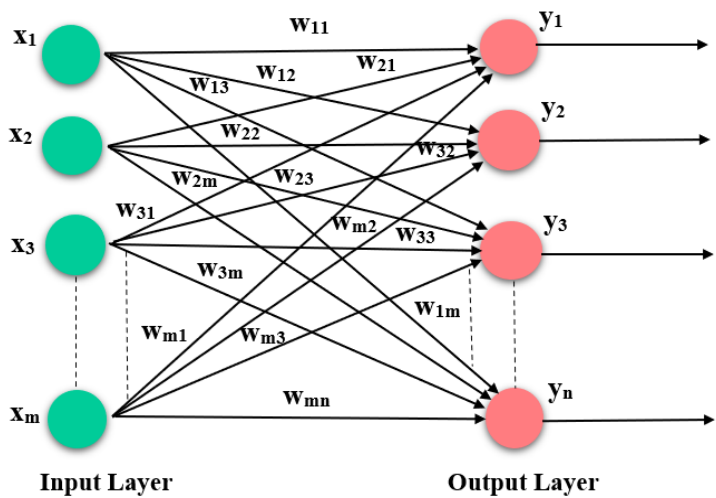

**Input Layer**        **Output Layer**

### 1. Functional Representation

For output neuron $y_1$:

$$y\_{in1} = x_1 w_{11} + x_2 w_{21} + \ldots + x_n w_{n1} = \sum_{i=1}^{n} x_i w_i \qquad (1)$$

In vector notation:

$$y\_{in1} = [x_1 \; x_2 \; \ldots \ldots \; x_n] \times \begin{bmatrix} w_{11} \\ w_{21} \\ w_{n1} \end{bmatrix} = X \times W_1$$

where $X = [x_1 \; x_2 \; \ldots \; x_n]$ is the input vector of the input signal, and $W_{*1}$ is the first column of the weight matrix as:

$$W = \begin{bmatrix} w_{11} & w_{12} & w_{1m} \\ w_{21} & w_{22} & w_{2m} \\ w_{n1} & w_{n2} & w_{nm} \end{bmatrix}$$

In general, the net input $y\_{inj}$ to the output unit $Y_j$ is given by:

$$y\_{inj} = [x_1 \; x_2 \; \ldots \ldots \; x_n] \times \begin{bmatrix} w_{1j} \\ w_{2j} \\ w_{nj} \end{bmatrix} = X \times W_{*j} \qquad (2)$$

If $Y\_{in}$ denotes the prime vector for all the net inputs to the array of input units,

$$\sum Y\_{in} = [y\_{in1} + y\_{in2} + \ldots \ldots + Y\_{in\;n}]$$

Then the next input to the entire array of output can be expressed concisely in matrix notation as:

$$Y\_{in} = X \times W \qquad (3)$$

The signals transmitted by the output units, depending on the nature of the concerned activation function (Karlik & Olgac, 2011).

## HEBBIAN LEARNING TECHNIQUE

A neural network can be trained or it can learn by several learning methods, such as Supervised Learning, Semi-Supervised Learning, Unsupervised Learning, Reinforced Learning, etc. The most commonly used learning methods are Supervised Learning and Unsupervised Learning, which can be further categorized (Karlik & Olgac, 2011). In the Supervised Learning method, every input pattern that is used to train the neural network is associated with an output pattern. On the other hand, in Unsupervised Learning, the output is not presented to the network, rather the network learns by its own by discovering to structural features of input patterns. The authors are considering the Hebbian Learning technique (Unsupervised Learning) featured with music feature labeling for the learning process of single-layer neural network.

Hebb Rule, or Hebbian Learning technique tautologize repeatedly in an input-output pattern, which can also be depicted as training vectors. In Hebbian Learning, if two cells fire simultaneously, then the strength of the connection or weight between them should be increased. The increment in weight between two neurons is proportional to the frequency at which they operate together. Moreover, as the Hebb Rule accepts input/output patterns in bipolar form, the distinct music features can be labeled as different binary values, and henceforth, the authors can proceed by creating a training set by using AND function.

First, they consider the general training pattern of a neural network with Hebb's Rule, then they proceed further with the case-specific training for their approach.

Let us assume that, we have a pair of training vectors α and β, in correspondence with the input vector h.

According to Hebb's rule, let us first assign all the initial weight to 0, i.e.:

$$W_{ij} = 0, \text{ where } i = 1\ldots.m \text{ and } j = 1\ldots\ldots.n \qquad (4)$$

For each input-output training epoch, the new input variable in the current input set will be:

$$\alpha_i = h_i \; (i = 1\ldots\ldots\ldots.m)$$

With correspondence, the new output variable to the current output set will be:

$$\beta_j = g_j \; (j = 1\ldots\ldots\ldots..n)$$

Henceforth, the authors have to adjust the weight using the following equation:

$W_{ij \text{ (new)}} = w_{ij \text{ (old)}} + h_i g_j$, $(i = 1\ldots m \ \& \ j = 1\ldots n)$ (5)

From the above-mentioned equation, they can set the activation units for the binary output results of the neural network:

$$g_j = 1, h > 0 \ (1)$$
$$= 0, h \leq 0 \ (-1)$$

Following is the training input table:

*Table 1. Training input table generated using the Hebbian learning algorithm*

| $h_0$ | $h_1$ | $h_2$ | Output |
|---|---|---|---|
| 1 | 1 | 1 | 1 |
| 1 | 1 | -1 | -1 |
| 1 | -1 | 1 | -1 |
| 1 | -1 | -1 | -1 |

During training, all weights are initialized to 0.

So, $w_0 = w_1 = w_2 = 0$

The initial equation for Heb Rule:

$$\Delta W_i = h_i \times t$$

Now, the revised equation:

$\Delta w_i \text{ (new)} = w_i \text{ (old)} + \Delta w_i$ (where, $\Delta w_i = w_i \times t$) (6)

In the succession of this equation, the authors present the following learning iterations in a tabular form:

*Table 2. Learning iterations with a complete input vector set*

| Training Patterns | $h_0$ | $h_1$ | $h_2$ | t | $\Delta w_0$ | $\Delta w_1$ | $\Delta w_2$ | $w_0$ | $w_1$ | $w_2$ |
|---|---|---|---|---|---|---|---|---|---|---|
|   |   |   |   |   |   |   |   | 0 | 0 | 0 |
| 1 | 1 | 1 | 1 | 1 | 1 | 1 | 1 | 1 | 1 | 1 |
| 2 | 1 | 1 | -1 | -1 | -1 | -1 | 1 | 0 | 0 | 2 |
| 3 | 1 | -1 | 1 | -1 | -1 | 1 | -1 | -1 | 1 | 1 |
| 4 | 1 | -1 | -1 | -1 | -1 | 1 | 1 | -2 | 2 | 2 |

## EXPERIMENTAL SETUP

### 1. Setup

They authors use Google Colab as their platform for experimentation. It is a cloud-based Python notebook platform primarily aimed at Python versions over 3. Also, it introduces a native feature named as a hardware accelerator for faster execution time, but with a limited resource threshold. Colab provides the options to choose from the dedicated graphical processor environment execution based on Tesla K80 GPUs or Google's tensor processing units, developed for parallel neural computations simultaneously. While users can select any of the previously mentioned options, the authors select none, because there is a significant delay in resource granting for adjusting usage limits and hardware availability if the GPU or TPU accelerator is selected. They also connect to a hosted runtime with sufficient but abstract RAM and disk availability, as Google does not share the exact figurative information with the users.

### 2. Hyperparameters Tuning

The authors have used 100 layers for a simple feedforward structure. The dense output layering holds the activation classifiers. They then initialize it with loss function to evaluate the training loss. The authors keep the learning rate of the optimizer as 0.01. The dropout rate for avoiding overfitting is respectively 0.6 for the vectorization. they fit the model within the padded vector-matrix in the x-axis and y-axis consecutively. They set the epochs for the neural network as 100. The verbose information is kept as 1 for word (vector) to training logs. Finally, the authors print the collective outcome as model.summary( ).

## TRAINING DETAILS

With the previously discussed model learning inputs, the authors also need to show the musical features in correspondence with a similar binary form, from which the neural network will learn and will be able to recognize such kind of features after several epochs. Hence, they explain the bipolar labeling of the different musical features as mentioned earlier. These features were initially proposed by G. Tzanetakis (2002). By following these features, (Li et al., 2003) made an accuracy table of different Machine Learning methods tested on the GTZAN dataset (Strum, 2013), which they have previously mentioned in section 4.

The authors take a reference to only four distinct musical and speech recognition features from their experiment trained by them using Support Vector Machine. These features are Beat, Fast Fourier Transform (FFT) which decomposes a signal into many signals in a particular given time point, Mel-Frequency Cepstral Coefficient (MFCC) which is a collection of visibly specific speech recognition features, and finally, Pitch. The table is represented below:

*Table 3. Certain musical features in Single and Multi-class objective functions*

| Features | Methods (Train) | | Methods (Test) |
|---|---|---|---|
|   | SVM1 | SVM2 | FFNN |
| Beat | 26.5 (3.30) | 21.5 (2.71) | 88.33% |
| FFT | 61.2 (6.74) | 61.8 (3.39) | 78.16% |
| MFCC | 58.4 (3.31) | 58.1 (4.72) | 70.29% |
| Pitch | 36.6 (2.95) | 33.6 (3.23) | **90.36%** |

Here SVM1 suggests single class objective function, and SVM2 suggests multi-class objective function. As the authors are using a single layered feedforward network for their approach, FFNN denotes their network's performance in detecting the features in the validation phase. They further take each feature's distinct value in SVM1 and convert them into corresponding binary values. Henceforth the authors get the following results:

*Table 4. Conversion of single-class objective function values into a corresponding binary representation*

| Features | Method |
|---|---|
|   | SVM1 |
| Beat | 00011010 |
| FFT | 00111101 |
| MFCC | 00111010 |
| Pitch | 00100100 |

Now, the authors can represent these distinct binary values for each feature as the input patterns (or input signals) of their neural network, as shown in Figure 4.

Further following these distinct input patterns for each musical feature, they developed the whole feedforward neural network for training it for the n number of iterations as per the 66% of the total dataset. Figure 5 represents the simulation diagram of their proposed feedforward network.

*Figure 4. Input patterns (or input signals) of perceptrons for different music features*

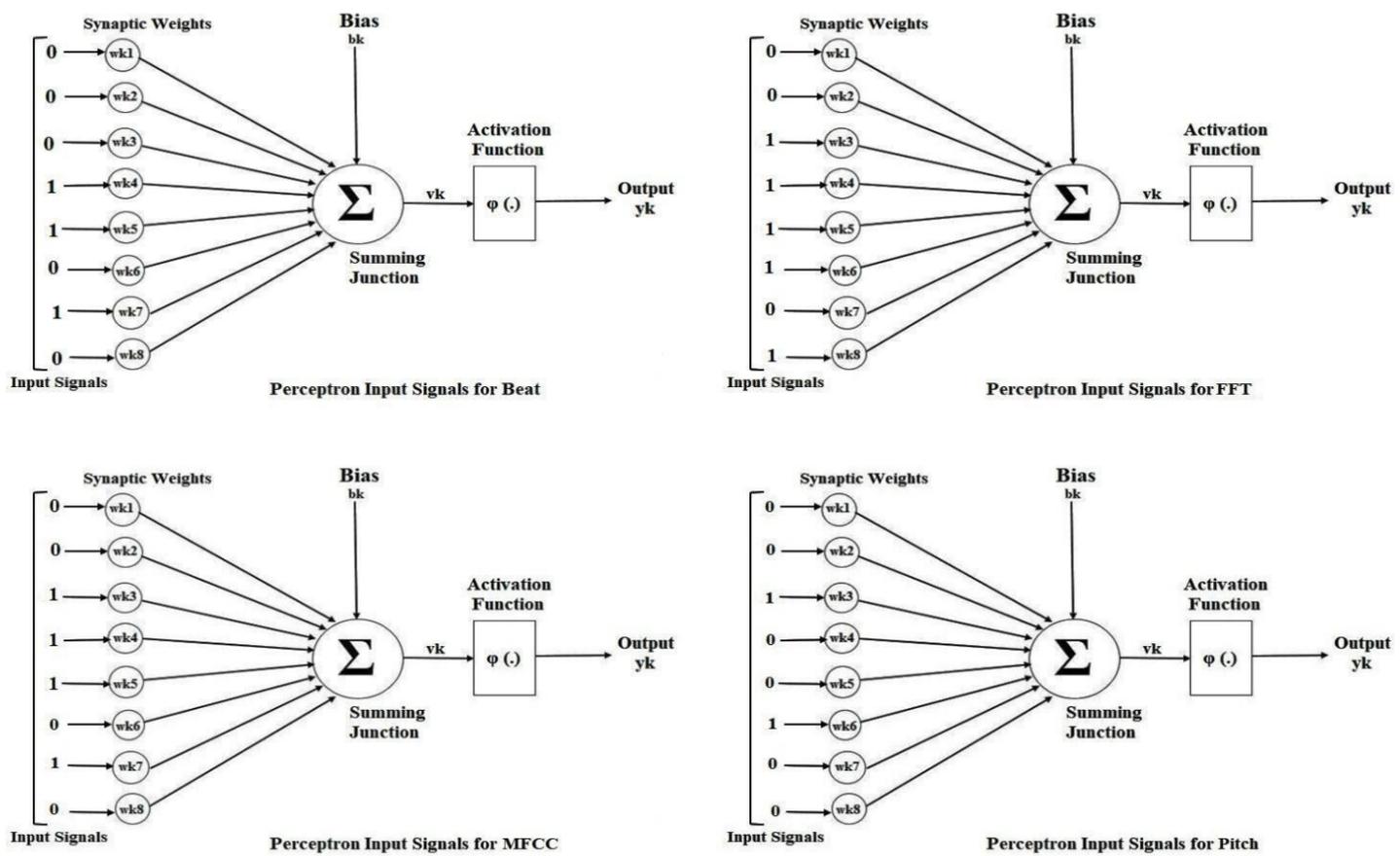

*Figure 5. Feedforward neural network model visualized for the experiment*

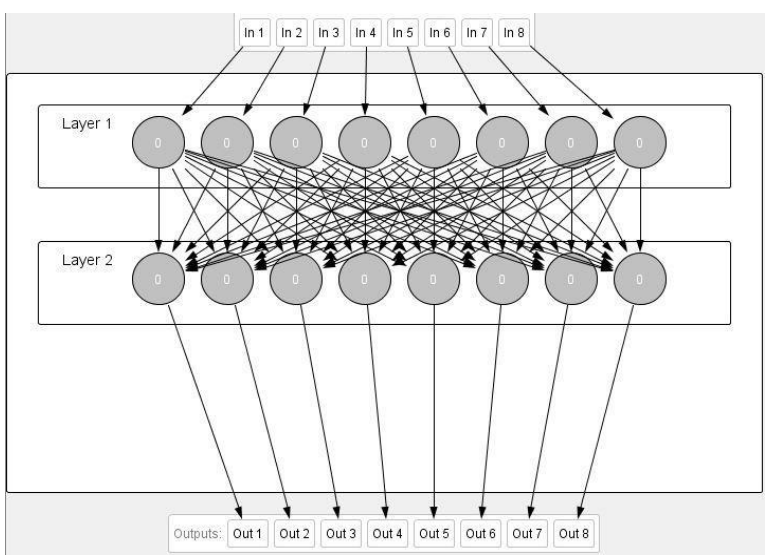

Once the neural network was developed, they started the unsupervised Hebbian technique training with the binary input values that the authors previously gained from the distinct musical features as mentioned earlier. The kept the maximum error rate as 0.01, learning rate ($\eta$) as 0.2 per second for providing a reduced number but healthy time for each epoch to train the neural network properly, and finally, they kept the momentum of learning as 0.7. Henceforth, they train their network for 10,000 iterations.

Proposed algorithm for training the neural network is as follows:

*Algorithm 1. Training procedure using Hebbian unsupervised algorithm.*

| Training Procedure with Hebbian Learning |
|---|
| 1. *music_feature_data* ← numpy.array |
| 2. ([ |
| 3.    [*Binary_Input_Dataset*]   //*input combinations* |
| 4. ]) |
| 5.   *feature_test_cases* ← numpy.array |
| 6.   ([ |
| 7.     [*Binary_Test_Dataset*] |
| 8.   ]) |
| 9.   *hebbnet* ← algorithms.HebbRule |
| 10.   ( |
| 11.     add *inputs* for *training* |
| 12.     add *outputs* for *testing* |
| 13.     def (*step*) |
| 14.     def (*learning_rate*) |
| 15.   ) |
| 16.   *hebbnet.train*(music_feature_data def (*epochs*)) |
| 17.   *hebbnet.predict*(feature_test_cases) |
| 18. *end* |

## RESULT & OBSERVATIONS

While training the neural network with the input vector sets as mentioned before, the ran a total network error graph parallelly to find out the cumulative occurrence of errors during the iteration time to time. The authors present this graph in Figure 6.

From Figure 6, the authors can visualize that with the increment of the iterations, the error rate in the network significantly reduced from 0.2 to almost 0.0. That depicts the learning ability and overall error reduction capability of their developed network with the iterations over time. With that, the authors also plotted the histogram graph of the weights of their musical features as input data.

Since they have 8 inputs altogether, for the convenience of fitting the data properly, they only show the histogram plotting of three Most Significant Bit's (MSB) graph from the left-hand side in figure 7. After the learning iteration is completed, the authors tested their network with the rest of the split validated dataset (34%), and they present the following results in Table 5 form as they encountered.

*Figure 6. The total error generated in the network during training iterations*

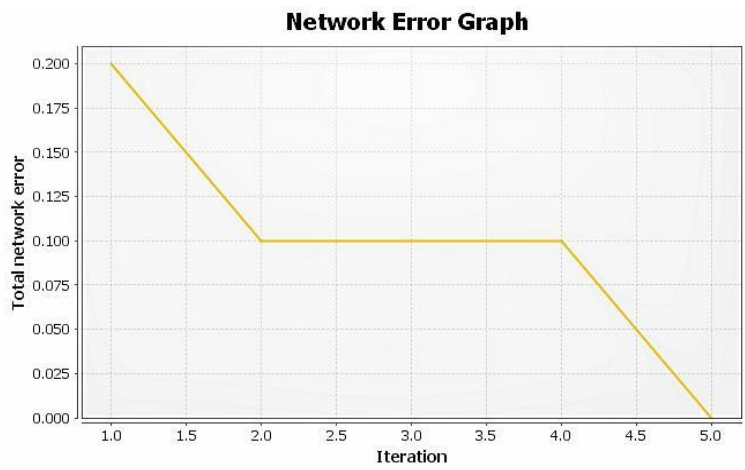

*Figure 7. Histogram representation of network input weights*

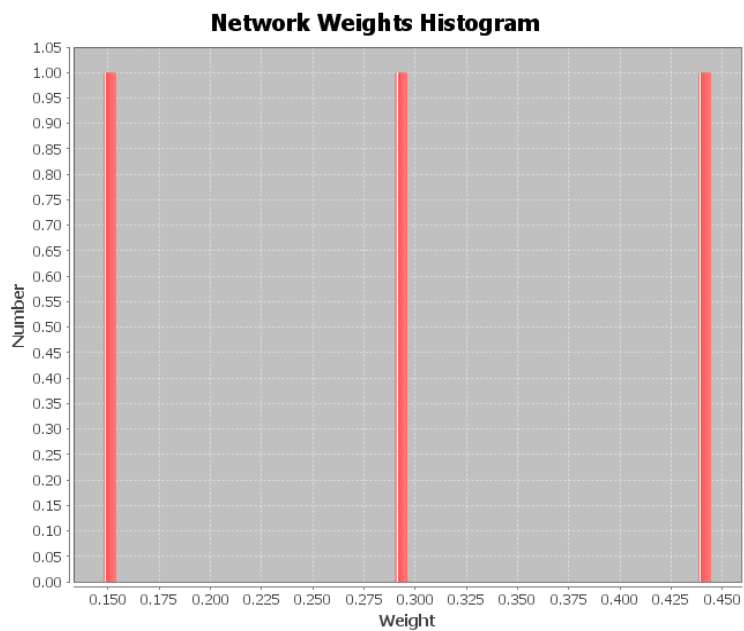

*Table 5. Representation of error rate, actual output, and desired output of each corresponding input*

| Input | Actual Output | Desired Output | Error |
|---|---|---|---|
| 1010101 | 1110001 | 1010001 | -100000 |
| 1010100 | 1010110 | 1010010 | -100 |
| 1010101 | 1010011 | 1010011 | 0 |
| 1010011 | 0010011 | 1010010 | 111111 |
| 1010010 | 1011011 | 1010011 | -1000 |
| 1010011 | 1110100 | 1010001 | -10011 |
| 1010010 | 1010010 | 1010010 | 0 |
| 1010001 | 1000101 | 1010101 | 100 |

The adaptive representation of the difference between actual outputs (ρ) and the desired outputs (κ), i.e. the Least Mean Square Error rate achieved by us was 1.021348 (correct up to 6 decimal places).

Hence, the authors can further represent the LMS as:

Desired Output (κ) - Actual Output (ρ) ≤ LMS Error

More formally,

$$\kappa - \rho \leq 1.021348 \quad (7)$$

It is worth mentioning that if the iteration is done for more stretch, i.e. for a larger number of epochs, the LMS rate will decrease and the learning accuracy will rise over time.

Furthermore, the authors plotted a comparative graph between the Actual Output, Desired Output, and Error from the values that they gained from Table 5, to study and visualize the data dependency and cohesiveness among the actual and desired output. The authors present this graph in Figure 8.

*Figure 8. The comparative graph between actual output, desired output, and error*

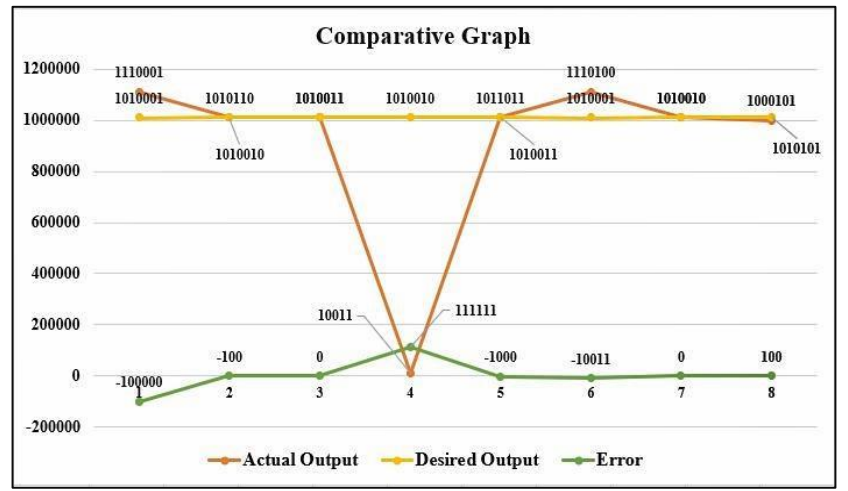

From the comparative graph, they can see that the difference between the actual output and the desired output is very nominal; in fact, they intercepted each other on several occasions, and there is only one visible case where the actual output performs poorly and hit the infra-index below of the error rate.

1. **Comparative Analysis**

For the equipollent comparison among similar tasks, the authors present the case comparisons as the direct accuracy comparisons with similar accuracy and/or F measures from previous experimentations with similar baseline deep neural network used. The authors' intention here is to scale comparative performance measurements by neural network models in similar large public datasets. Table 6 shows various models in comparison based on public music datasets.

(Grill & Schluter, 2015) combined the spectrograms with convolutional networks for long term self-similarity relations among different music boundaries. Among the feature combinations, they achieved a peak F1 score of 52.30%, resulting in new state-of-the-art output. (Dai et. Al., 2016) proposed the music feature segmentation using a long-short term network. They trained their model on sequential music features with softmax output classifier, and then segmentation was used to distinguish among such features. By this approach, they notice an improved classification accuracy of 89.71%. (Costa et al., 2017) deployed a convolutional network to classify music genres. The researchers proposed an approach for checking whether the music features learned from training are complementary or not, concerning time-frequency. They further tuned and com-pared among several parameters with Boltzmann pattern. Their combination of a convolutional network with Boltzmann feature learning pattern achieved an overall accuracy of 92% on public data. (Schreiber & Muller, 2018) presented deployed a convolutional neural network for musical tempo estimation. The networks trained on data augmented music corpus consisting of a vast variety of genres. Instead of observing the full tracks, they observed the behaviors from short pieces. Finally, they approached for generating the mean from all the gathered tempo information and made a global tempo. (Solanki & Pandey, 2019) used a similar neural network model, but they modified a traditional CNN into a deep network with more hidden layers. The network was trained on fixed-length polyphonic music with variable audio lengths. They used max-pooling layers as pre-output classifiers to reduce the dimensionality, and softmax as the output classification function. They reported achieving 92.80%. overall accuracy.

*Table 6. Comparison of the proposed work with similar tasks for music feature detection*

| Model | Data | Accuracy (%) | F1-Score (%) |
|---|---|---|---|
| CNN+Spectogram (Grill & Schluter, 2015) | SALAMI | - | 52.30 |
| LSTM+Segmentation (Dai et. Al., 2016) | ISMIR | 89.71 | - |
| SSD+CNN+RLBP (Costa et al., 2017) | ISMIR | 86.70 | 86.60 |
| CNN+Tempo (Schreiber & Muller, 2018) | GiantSteps | 89.30 | - |
| CNN+Max Pooling (Solanki & Pandey, 2019) | IMRAS | 92.80 | - |
| FFNN+HebbNet (proposed) | GTZAN | **90.36** | **89.27** |

As depicted from Table 6, the proposed method for music feature detection outperforms several previous benchmark results, while narrowly falling short to that of (Solanki & Pandey, 2019). Simultaneously, the authors' proposed network maintains good linearity between both the overall accuracy and F measure (F1-score).

## RUNTIME COMPLEXITY

The runtime complexities of the proposed neural networks are associated with all possible scenarios of space and time that are needed to train and validate such a network. The runtime complexity can be calculated from two distinct perspectives within the experimented neural networks:

1. There are two different complexities for different operational neural networks. For instance, a network containing backpropagation has to be calculated the complexity of backpropagation passes along with the feedforward cycles. These two sets of complexities are then summarized and made mean to represent the optimal complexity. However, as the authors deal with only a feedforward network here, they omit the latter part from consideration.
2. Also, the complexity is distributed both in the training and testing phases of the neural network. Both phases can be represented together for inferring the local optimum system, where any of the phases is highly complexity efficient, but the other phase is not. Similarly, the global optimum system can also be represented, where both the test and validation phases comprise of highly minimal computational complexities.

For forward propagation, the authors look for the asymptotic complexity of the pass. They assume the input vector generated for training is:

$$i \in \mathbb{R}^n$$

Here the first bias element is $i_0 = 1$, and $\mathbb{R}$ determines the rank of the sequence.

Now for the activation function cumulation, the index will start from 0, and the functions will be indexed in a matrices form. As the authors have shown previously, the value of the 0th element is 1.

To evaluate each activation value as an element of the matrix, they can deduct the following:

$$a^k = g(f^k)$$

where, f represents the feedforward pass, and: $f^k = i^{n-1}$

For each layer of such matrix element computation, an activation function generates a value output. Now, from the matrix computations of its elements, the asymptotic complexity is $O(n^3)$. Accordingly, since f is the basic representation of a forward activation, and there is n number of others, hence f has a runtime of $O(n)$. Also, in the learning phase, the runtime complexity of the forward pass is faster than that of the back-propagation pass; resulting in a non-achievable optimum complexity for the networks having a larger number of hidden layers. Consecutively, the training and test time of binary classification problems are lower than multitask classifications.

## DISCUSSION

From the representation of their work, the authors saw a simple FFNN can produce an impressive accuracy for feature detection. This enhancement comes from the unsupervised learning algorithm, which provided the network with better learning cycles (epochs). Also, instead of discussing the practical details of the programming side, they are rather keen to explore the empirical side of the proposed network on how it can be fine-tuned for the best possible music feature recognition following their approach. The authors have also depicted the comparisons and error rates to understand where does their proposed model stands in comparison with the previous gold standard works.

1. **Future Scopes**

As much as works have been carried forward for music feature identification using various neural network models, not many works have explored the empirical details of how the layers work during training and test phases during such experimentations. Also, no song retains the same pitch, tempo, and frequency for the entire playtime. It changes from low to high and vice versa with time. Hence it is important to exploit the seg-mentation methods to make small but cohesive segments of a track concerning time, for better understanding of the vivid characteristic changeover of any track. As the authors have represented how an unsupervised training can affect the expressional outputs from bias junctions of a set of neuron setups, these settings are yet to be explored for deep neural networks (DNN). Finally, instead of binary classifications, a better understanding of music features comes with multiclass classification consisting of ensemble feature combinations. This is an area authors believe holds high potential for future scopes.

## CONCLUSION

In this paper, the authors represented an experimental work of distinct and versatile music features identification using a single layer feedforward neural network with the unsupervised algorithm. The authors proposed the training mechanism of the neural network, and the different musical features as the inputs of the neural network, and implemented it in this task. However, for future work, they want to carry this idea as an extension work on a more complicated network, i.e. Multi-Layer Feedforward Network (Back-propagation Network) with real-time visual and performance simulation. Alongside that, they also intend to test and compare the performance of several Supervised and Unsupervised neural network learning techniques as an expansion for this particular task. Besides, they will study and gather an extensive and vast musical feature set from vivid and versatile music genres like Rock, Pop, Jazz, Western Classical, etc. Once the dataset is developed, it will allow us to look into more intricate details for sophisticated music genre identification and portray the results.